# A spatiotemporal style transfer algorithm for dynamic visual stimulus generation

Antonino Greco[1,2,3,*] and Markus Siegel[1,2,3,4,*]

[1] Department of Neural Dynamics and Magnetoencephalography, Hertie Institute for Clinical Brain Research, University of Tübingen, Germany
[2] Centre for Integrative Neuroscience, University of Tübingen, Germany
[3] MEG Center, University of Tübingen, Germany
[4] German Center for Mental Health (DZPG), Tübingen, Germany
[*] Corresponding authors: Markus Siegel (markus.siegel@uni-tuebingen.de) and Antonino Greco (antonino.greco@uni-tuebingen.de)

**Understanding how visual information is encoded in biological and artificial systems often requires vision scientists to generate appropriate stimuli to test specific hypotheses. Although deep neural network models have revolutionized the field of image generation with methods such as image style transfer, available methods for video generation are scarce. Here, we introduce the Spatiotemporal Style Transfer (STST) algorithm, a dynamic visual stimulus generation framework that allows powerful manipulation and synthesis of video stimuli for vision research. It is based on a two-stream deep neural network model that factorizes spatial and temporal features to generate dynamic visual stimuli whose model layer activations are matched to those of input videos. As an example, we show that our algorithm enables the generation of "model metamers", dynamic stimuli whose layer activations within our two-stream model are matched to those of natural videos. We show that these generated stimuli match the low-level spatiotemporal features of their natural counterparts but lack their high-level semantic features, making it a powerful paradigm to study object recognition. Late layer activations in deep vision models exhibited a lower similarity between natural and metameric stimuli compared to early layers, confirming the lack of high-level information in the generated stimuli. Finally, we use our generated stimuli to probe the representational capabilities of predictive coding deep networks. These results showcase potential applications of our algorithm as a versatile tool for dynamic stimulus generation in vision science.**

## Introduction

Understanding how visual information is encoded in the brain has been a longstanding goal of neuroscience and vision science. Recently, there has been an increasing interest among vision scientist also to study hidden representations of computer vision models and to compare biological and artificial vision across all levels of analysis [1]. A fundamental component of this research is the generation of controlled visual stimuli. Traditionally, stimulus manipulation in vision research has been performed for low-level features of static visual stimuli such as the matching of pixel contrast [2], or manipulating the speed of dynamic visual stimuli [3,4]. Recently, computer graphics approaches led to a proliferation of methods that can generate parametrically designed images both machine learning and neuroscience research. Examples include the parametric generation of face expressions [5] or 3D visual scenes [6,7].

While these methods have significantly progressed the field of stimulus generation, they often fall short in terms of flexibility and tend to diverge from the natural statistics of our visual environment [8], frequently resulting in outputs that can appear artificial. In contrast, deep neural network models (DNN) trained for computer vision tasks [9,10] have revolutionized the approach of image synthesis in a variety of ways, affording researchers novel paradigms to investigate visual processing within artificial and biological neural networks [11–13]. One of the earliest examples of DNN-based stimulus generation method is DeepDream [14–18], an algorithm that accentuates patterns in images in a manner that can be conceived as "algorithmic pareidolia", using a pre-trained deep convolutional neural network (CNN) [19]. This method has been predominantly used in vision research to simulate the visual hallucination patterns commonly found in altered states of consciousness induced by psychedelic drugs [20–22]. An alternative to this method has been also proposed to generate visual stimuli that maximally activate specific visual cortical regions across species [13,23,24]. Another influential family of DNN-based methods for image synthesis comes from the seminal work of Gatys and colleagues [25], in which they proposed an algorithm, namely the Neural Style Transfer (NST), for extracting and recombining the texture and shape of natural images to generate novel stimuli [25–30]. NST has been applied in vision research to compare visual representations of images in humans and machines [31–34] and to understand how spatial texture information is encoded in the visual cortex of the mammalian brain [35–40].

Despite the abundance of methods for generating static visual stimuli, the algorithms proposed for dynamic visual stimuli generation (i.e., videos) are scarce. Unlike images, videos also have a temporal dimension and, in natural videos, the relationship between spatial and temporal features is often non-trivial [35,41,42]. Here, we introduce a method for generating dynamic visual stimuli that can serve different purposes in vision research. The method is based on the NST algorithm but extends its capabilities to videos. Thus, we refer to it as Spatiotemporal Style Transfer (STST).

We capitalized on recent texturization algorithms that extended the NST capabilities to generate dynamic textures (i.e., video clips with stationary spatiotemporal patters such as flames or sea waves) [43,44]. The general idea is to synthesize "model metamers" [45], which are dynamic stimuli whose layer activations within a DNN model are matched to those of natural videos. To create these stimuli, we perform optimization in a two-stream model architecture [42,44,45]. This model is designed to replicate the relative segregation of neural pathways responsible for processing spatial and temporal features in the brain, known as ventral and dorsal streams, respectively [48–51].

After elucidating the procedural steps of our algorithm, we demonstrate an example application in which we create model metamers starting from natural videos. The generated metamers exhibit matched low-level spatiotemporal features with respect to their natural counterparts but lack high-level features like object-level semantics. This approach can be used to study object recognition in both biological and artificial visual systems [52–56]. To test this, we employed different state-of-the-art vision models and tested if early and late layer activations differed between natural and metameric stimuli. Finally, employed the generated stimuli to gain insights into the internal representations of predictive coding networks performing next frame prediction.

## Results

### Model architecture

Our proposed algorithm is based on a two-stream model architecture, consisting of one module (or "stream") acting as a spatial feature detector on each frame of the target video, and another module performing feature detection in the temporal domain across consecutive pairs of frames. The model is agnostic to the specific implementations of both spatial and temporal modules, thus any pretrained differentiable model can be used. Thus, in the following we formally denote the spatial module as a differentiable function $S(x_t)$ that takes as input the current frame defined as a 3D tensor $x_t \in \mathbb{R}^{H \times W \times C}$, where $H$ and $W$ are the height and width of the image in pixels, $C$ is the number of channels in the frame (in our case $C=3$, as we are working with RGB-encoded frames) and the subscript $t$ denotes the time step. Each element of the tensor $x_t$, represents the intensity of the pixel at position $(H, W)$ in channel $C$, with a value in the range [0,1]. We also denote the temporal module as a differentiable function $\mathcal{T}(x_t, x_{t-1})$ that takes as input both the current ($x_t$) and previous ($x_{t-1}$) frame.

We opted to maintain consistency with previous works on NST and its extension to dynamic textures. Thus, we adopted VGG-19 [57] as the spatial module and the multiscale spacetime-oriented energy model (MSOE) [44,58] as the temporal module. VGG-19 is a pretrained CNN model [57] that was trained on the ImageNet1k dataset [59,60] for object recognition and it was used originally both in Gatys et al. [25] and in Tesfaldet et al. [44]. We also opted for this model since recent evidence showed that this type of architecture outperforms many other models in the task of style transfer or texturization [61], although the reasons for this remain debated [61–

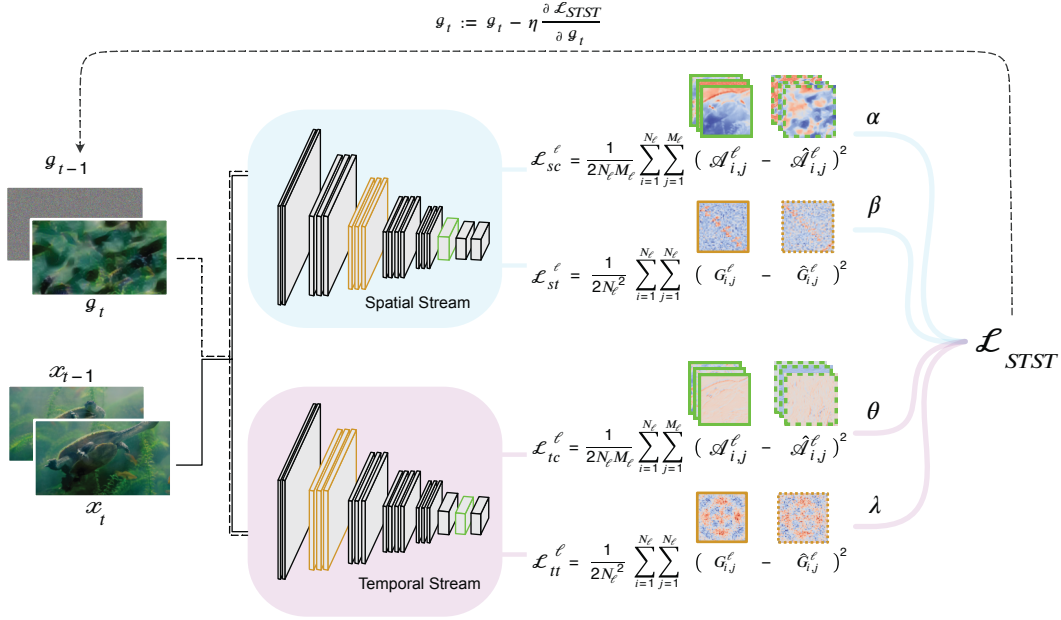

**Figure 1: Graphical representation of the STST algorithm.** The current frame as well as the current and previous frames are forward passed to the spatial and temporal stream, respectively, up to a predefined set of layers. A content and texture loss are defined for both streams as the difference of layer activations and their gram matrices between the target and generated frames, respectively. These are weighed and summed to form the total STST loss. The partial derivative of this loss with respect to the current generated frame is used to optimize it by gradient descent.

63]. MSOE is a pretrained CNN model that was trained on the UCF101 dataset [64] to predict optical flow and was similarly used in Tesfaldet et al. [44] as the temporal module. We opted for this model because it is generally invariant to spatial features [44] and is thus relying entirely on the temporal dynamics of the video data to estimate optical flow.

We denoted the layer activations of the CNN models as $\mathcal{A}^\ell \in \mathbb{R}^{N_\ell \times M_\ell}$, where $N_\ell$ and $M_\ell$ refer to the number of filters and the number of spatial locations (the height times the width of the feature map) in the convolutional layer $\ell$. The layer activation $\mathcal{A}^\ell$ is defined as the results of the forward pass to that specific layer $\ell$ after both the convolutional operator and the nonlinear activation function. Note that, here we refer to the layer activation $\mathcal{A}^\ell$ for both the forward pass of $\mathcal{S}(x_t)$ and $\mathcal{T}(x_t, x_{t-1})$, indiscriminately.

**Optimization procedure**

We now define the optimization procedure to generate what we referred to as "model metamers", which are generated dynamic stimuli possessing certain spatiotemporal features that are indistinguishable from their natural counterparts for the two-stream model [45]. The STST algorithm can generate metamers that are matched in terms of content (shape) and style (texture) along both spatial and temporal streams. Following the work of Gatys et al. [25], we define the loss function of the content matching procedure as the squared difference between the layer activations $\mathcal{A}^\ell$ of the target frame $x_t$ and the layer activations $\hat{\mathcal{A}}^\ell$ of the generated frame $g_t \in \mathbb{R}^{H \times W \times C}$:

$$\mathcal{L}^\ell_{content}(x_t, g_t) = \frac{1}{2N_\ell M_\ell} \sum_{i=1}^{N_\ell} \sum_{j=1}^{M_\ell} (\mathcal{A}^\ell_{i,j} - \hat{\mathcal{A}}^\ell_{i,j})^2 \qquad (1)$$

Optimizing this loss function for all frames allows the algorithm to match the spatial or temporal structure of the target video. In other words, by using merely the content matching procedure the algorithm generates approximately the same video that was used as target. Note that, for the temporal module $\mathcal{T}(x_t, x_{t-1})$ the input is both the current $x_t$ and previous $x_{t-1}$ target frame, thus also the current generated frame $g_t$ and the previous generated frame $g_{t-1}$ need to be provided.

Conversely, to perform texture matching we need to first obtain a representation of texture information [25]. Here, we refer to texture as either spatial or temporal recurrent patterns (i.e., a set of stationary statistics). We defined this texture representation as the correlation between different filter responses within a layer activation [25,44], encapsulated by the so-called Gram matrix $G^\ell \in \mathbb{R}^{N_\ell \times N_\ell}$, whose entries are given by:

$$G_{i,j}^\ell = \frac{1}{N_\ell M_\ell} \sum_{k=1}^{M_\ell} \mathcal{A}_{i,k}^\ell \, \mathcal{A}_{j,k}^\ell \qquad (2)$$

Thus, the loss function of the texture matching procedure is defined as:

$$\mathcal{L}_{texture}^\ell(x_t, g_t) = \frac{1}{2N_\ell^2} \sum_{i=1}^{N_\ell} \sum_{j=1}^{N_\ell} \left( G_{i,j}^\ell - \hat{G}_{i,j}^\ell \right)^2 \qquad (3)$$

where the $G^\ell$ is the Gram matrix computed from the layer activations $\mathcal{A}^\ell$ and $\hat{G}^\ell$ is computed from $\hat{\mathcal{A}}^\ell$. Generating metamers with the $\mathcal{L}_{texture}^\ell$ loss results in videos that no longer maintain a retinotopic correspondence with the original natural video. However, a variety of low-level statistical properties are preserved, depending on which dimension it is applied to. For example, in the spatial stream the luminosity and contrast are well preserved, as well as the magnitude of the optical flow or the global motion in the temporal stream. We can finally define the loss functions for the spatial and temporal stream as:

$$\mathcal{L}_{spatial}(x_t, g_t) = \sum_{\ell=1}^{L_s} \left( \alpha \mathcal{L}_{sc}^\ell(x_t, g_t) + \beta \mathcal{L}_{st}^\ell(x_t, g_t) \right) \qquad (4)$$

$$\mathcal{L}_{temporal}(x_t, g_t, x_{t-1}, g_{t-1}) = \sum_{\ell=1}^{L_t} \left( \theta \mathcal{L}_{tc}^\ell(x_t, g_t, x_{t-1}, g_{t-1}) + \lambda \mathcal{L}_{tt}^\ell(x_t, g_t, x_{t-1}, g_{t-1}) \right) \qquad (5)$$

and the general form of the loss function for our STST algorithm:

$$\mathcal{L}_{STST}(x_t, g_t, x_{t-1}, g_{t-1}) = \mathcal{L}_{spatial}(x_t, g_t) + \mathcal{L}_{temporal}(x_t, g_t, x_{t-1}, g_{t-1}) \qquad (6)$$

where $L_s$ and $L_t$ are the number of selected layers in the spatial and temporal modules, respectively. Moreover, $\alpha$ and $\beta$ represent the weights associated with the content $\mathcal{L}_{sc}^\ell$ and texture $\mathcal{L}_{st}^\ell$ loss in the spatial loss $\mathcal{L}_{spatial}$, as well as $\theta$ and $\lambda$ are the weights for the content $\mathcal{L}_{tc}^\ell$ and texture $\mathcal{L}_{tt}^\ell$ loss in the temporal loss $\mathcal{L}_{temporal}$. These weights are set as hyperparameters that modulate the relative contribution of each of the four terms to the general loss function $\mathcal{L}_{STST}$. Furthermore, the selection of a restricted number of terms (e.g., selecting only the spatial and temporal texture losses) can be seen as setting the undesired terms' weights to zero. This provides a versatile tool for dynamic stimulus generation, since one can perform many combinations of the four losses that compose the general loss function. Crucially, one can select more than one target video and even assign a different target video for each of the four loss terms.

Once the total loss for the optimization procedure is defined, the actual update of the parameter space $g_t$ consists of applying a gradient-based optimization method such as gradient descent, which iteratively adjusts the parameters in the direction that minimizes the loss:

$$g_t := g_t - \eta \frac{\partial \mathcal{L}_{STST}}{\partial g_t} \qquad (7)$$

where $\eta$ is the learning rate that weight the gradient of the loss with respect to the parameters and is set as an hyperparameter. The number of iterations needed to update the parameters $g_t$ is also an hyperparameter of STST. Importantly, the gradients are computed and applied only with respect to the current generated frame, even in the case in which the temporal module is performing the forward pass with both current and previous frames. This is because optimizing the full pair of consecutive frames would lead to instabilities in the optimization. The generation of the dynamic stimulus is finished when the optimization procedure has been iterated across all frames.

**Perceptual stabilization via preconditioning**

Although the optimization procedure described above performs well for videos with simple spatiotemporal dynamics, as similarly shown by Tesfaldet et al. [44] for the generation of dynamic textures, it encounters significant limitations when applied to natural video sequences. Empirically, it tends to manifest considerable perceptual instabilities both in space and time domain. Thus, we incorporated a family of techniques for image and video synthesis that significantly improved its robustness and perceptual stability. The employed techniques can be generally described as a "preconditioning" of the optimization procedure. In mathematical optimization, preconditioning refers to a transformation of an optimization problem to condition it to a form that is more suitable for finding optimal solutions. Intuitively, preconditioning changes the basins of attraction, allowing the optimization process to approach preferred solutions more easily [62].

The first preconditioning technique that we introduced is the addition of the total variation (TV) loss [65] to our loss function. The TV loss is a regularization term often used in image processing algorithms to encourage smoothness in the output while preserving edges, introduced in the literature of image generation from early works on the DeepDream algorithm [18,62]. Intuitively, the idea is that in natural images, pixel values tend to change gradually except at edges where there are abrupt changes. The total variation loss penalizes the sum of the absolute differences between neighboring pixel values, as described by the following anisotropic 2D version of the formula:

$$\mathcal{L}_{TV}(g_t) = \frac{1}{HWC} \sum_{k=1}^{C} \sum_{i,j}^{H,W} |g_t^{i+1,j,k} - g_t^{i,j,k}| + |g_t^{i,j+1,k} - g_t^{i,j,k}| \quad (8)$$

This additional loss helps to prevent the excessing high-frequency noise that results from optimization procedure described above. Specifically, this regards the spatial module and most probably the convolutional nature of the selected model, since it has been shown that strided convolutions and pooling operations can create high-frequency patterns in the gradients [62,63]. Note that this loss term is applied on the current generated frame during the optimization process, thus there is an additional hyperparameter $\omega$ to weight the contribution of the $\mathcal{L}_{TV}$ term to the total loss.

Another technique we borrowed from the DeepDream field is the spatial multiscale approach [14,62]. It basically consists of starting the optimization with an image at a lower resolution and iteratively enlarging it to the original size. This is done to inject spatial frequency biases in the parameter space, by initially providing low-frequency patterns and then refining high-frequency ones. We implemented this approach by resizing both the targets and generated frames using bilinear interpolation and an hyperparameter $\sigma$ to control the so-called octave scale. The octave values $o$ are the exponent power of the octave scale used for multiplying them with the original frame size $H \times W$, such as $H_\sigma = H\sigma^o$ and $W_\sigma = W\sigma^o$. Note that $o = 0$ yields the original resolution. The aforementioned optimization procedure is applied on each octave, possibly even with different hyperparameters, such as e.g., different learning rates for different octaves. We followed the authors that introduced the multiscale approach and normalized the gradients of the total loss with respect to $g_t$ by dividing them with their standard deviation across the height $H_\sigma$ and width $W_\sigma$ dimensions [14,62].

We also implemented in the STST algorithm a method to match the color distribution between $x_t$ and $g_t$. We noticed that this substantially improved the similarity between the targeted and generated color distributions compared to only applying the optimization. Thus, following the multiscale procedure, we applied a color transfer algorithm as a postprocess step [66,67]. Instead of using a naïve approach such as histogram matching for each channel individually [68], the employed algorithm finds a transformation from the full 3 dimensional probability density function of the pixel intensity of the target frame to the generated frame and also reduces the grain artefacts by preserving the gradient field of the target frame [67].

Finally, we applied a blending operation on the frame transitions to reduce flickering artefacts. This turned out to be a critical step that improved perceptual stability by preconditioning the initial conditions of the optimization procedure at the current frame with the post-processed frames at the previous time point and the addition of uniform noise $\mu \in \mathbb{R}^{H \times W \times C}$, as following:

$$g_t \leftarrow \varphi g_{t-1} + (1-\varphi)\mu, \quad \mu \sim \mathcal{U}(0,1) \tag{9}$$

where $\varphi$ is the hyperparameter that controls the blending ratio and ← stands for initialization. For instance, setting $\varphi = 0.9$ results in the current frame being initialized with 90% the pixel intensity of the previous frame and the rest being uniform noise. Although this blending operation significantly improved the temporal stability of our method, initial frames of the generated stimuli had slightly different low-level statistics compared to the rest of the frames. In other words, the relative stationarity of the low-level statistics in the generated stimuli with respect to the target stimuli started to be consistent only after a few initial movie frames. We assumed that this was since the very first frame is initialized only as noise, because there is no previous frame to precondition with. Also, it is difficult to increase the speed of the optimization process as our algorithm needs to have a relatively low value of $\eta$ to maintain numerical stability. This can be interpreted as the search for stable local minima not just within each frame, but across them, as the parameter space of the next frames is preconditioned on the previous ones by the blending operation. Thus, to remedy the first frame effect, we mirror padded the target and the generated stimuli for the first $\xi$ frames. In other words, we took the first $\xi$ frames of the target stimuli, flipped their order and concatenate them before the first frame. We then discarded the initial padded frames after the optimization.

### Model metamers for object recognition

We now describe an example application of STST for investigating object recognition in visual systems. We generated metamer dynamic stimuli that had similar spatiotemporal low-level features compared to their natural counterparts but lack high-level object information. We collected 3 high-quality video clips from the YouTube-8M dataset [69] and optimized dynamic stimuli to match both their spatial and temporal texture. For comparison, we generated dynamic stimuli starting from the same target videos but using another algorithm [70] that was previously proposed to generate dynamic metamers [45]. This alternative algorithm basically consists of randomizing the phase in the spatiotemporal frequency domain [70]. In the following, we will refer to this alternative algorithm as Spatiotemporal Phase Scrambling (STPS).

By qualitatively inspecting the generated frames in Fig. 2a, it can be noticed how our STST algorithm and STPS produce very dissimilar stimuli both in terms of appearance and motion dynamics. We computed 4 basic spatiotemporal features from the original movies, our STST stimuli and the STPS stimuli to quantitatively assess both methods, namely pixel intensity and contrast as spatial features as well as pixel change and optical flow (magnitude and angle) as temporal features (Fig. 2b). We found that both methods excelled at matching spatial features, while temporal features were less preserved. Critically, STST performed substantially better than STPS at matching the optical flow of the natural videos, which is a crucial aspect for dynamic stimuli. Thus, these results showed how STST preserves spatiotemporal low-level statistics while high-level information is dismantled.

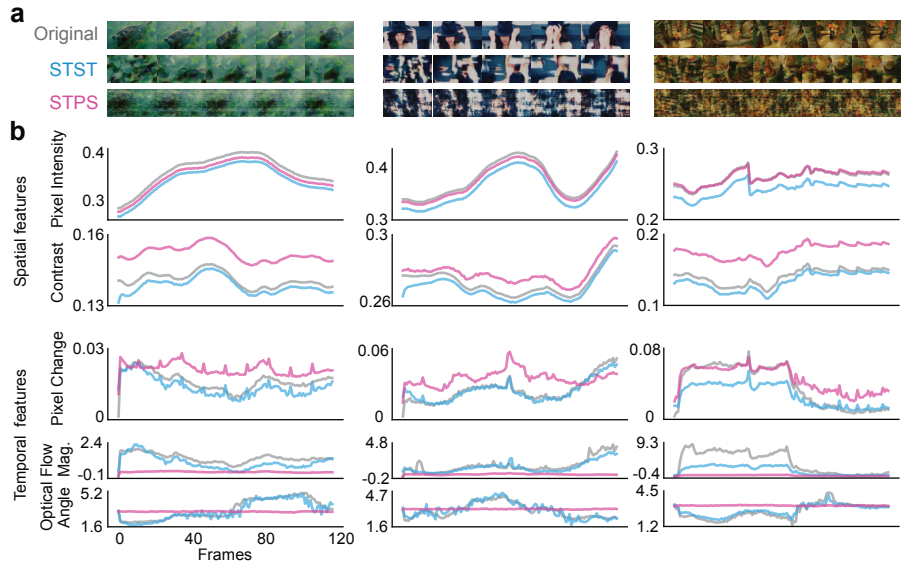

Figure 2: Spatiotemporal low-level features are preserved in STST stimuli. **a**, Example frames from the original target videos as well as from STST and STPS generated videos. **b**, Time courses of 4 spatiotemporal features in original (gray), STST (azure) and STPS (pink) stimuli. Features are pixel intensity and contrast as spatial low-level features and pixel change and optical flow (magnitude and angle) as temporal low-level features.

**Metamers disrupt late layer representations of vision models**

Next, we tested the generated stimuli using state of the art deep learning models for image and video classification. We used these models since they exhibit a hierarchical structure for extracting visual features similar to the mammalian visual cortex [9,71]. Thus, we hypothesized that the similarity between natural and STST stimuli in the layer activations of these models should be higher in early layers as compared to late layers (Fig. 3a). To compute the similarity between layers' activations, we used the Center Kernel Alignment (CKA) score [72]. Notably, we did not use the same models we deployed for the spatial and temporal stream in the STST model.

We found almost perfect matching of early layer representations between natural and metamer stimuli across all example videos in image classification models (Fig. 3b), which were ResNet50 [73] and ConvNeXt-T [74]. Conversely, late layer activations differed substantially between natural and metamer stimuli. This pattern of decreased similarity was consistent across all videos. Although these results were in line with our predictions, the employed models were applied on a frame-by-frame basis as they were conceived for image classification. Thus, we extended our analyses to video classification models using ResNet18-3D [75] and ResNet18-MC3 [76]. We employed the same procedure but this time using snippets of videos in a sliding-window approach. Again, we observed early layer activations being very similar across all frames and example videos, while late layers showed consistently decreased similarity. Crucially, we found the same pattern as in image classification models, although the range of the effects was shorter (Fig. 3c). Together, these findings confirmed our expectations that STST generates dynamic stimuli with high similarity to natural videos in terms of low-level spatiotemporal statistics but without preserving high-level features that are critical for object recognition in biological and artificial vision systems.

**Metamers suggest no semantic understanding in PredNet**

Finally, we used stimuli generated with STST to investigate the representational capabilities of deep neural networks inspired by predictive coding theories [77,78]. We focused our investigation on PredNet [79], a deep convolutional recurrent neural network (Fig. 4a) that was trained with self-supervised learning to perform next frame prediction [80]. We opted for this model since it possess interesting properties that are aligned with a range of phenomena observed in visual cortex [81], probably due to its architectural structure that involves recurrent and feedback connections [79] which are absent in traditional feedforward-only CNN models [56,82]. Recently, it has been debated if PredNet is able to extract robust high-level representations of its dynamic inputs or if it merely acts as a flow filter predicting low-level optical flow [77,83].

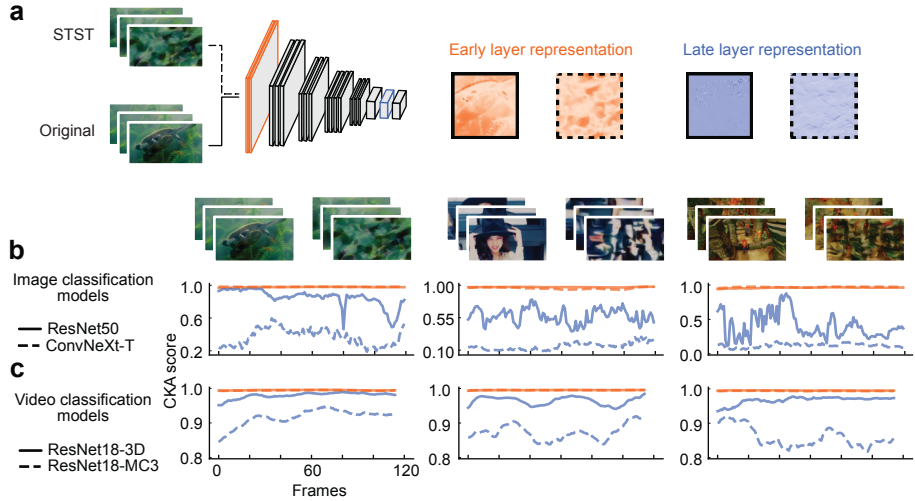

**Figure 3: Model metamers disrupt late layer activations in later stages of deep vision models. a**, we used STST stimuli and their natural counterparts as inputs to deep vision models and extracted early (orange) and late (blue) layer activations. **b**, time courses of the similarity metric (CKA) between natural and generated stimuli for image classification models. **c**, time courses of the similarity metric (CKA) for video classification models.

To test this, we compared next frame predictions PredNet for both original and STST generated metamer stimuli (Fig. 4a). We reasoned that next frame predictions should be better for natural as compared to metamere stimuli if PredNet was able to exploit high-level visual representations. Fig. 4b shows exemplary frame precitions by the PredNet model. Qualitatively, frame predictions had a high fidelity compared to their ground truth for both natural and metamer stimuli. We computed the structural similarity index measure (SSIM) [84] and its dynamic version conditional on the previous frame (cSSIM) [83] to quantitatively measure how well PredNet performed next frame prediction (Fig. 4c). We found that SSIMs were similar for natural and STST stimuli across frames and on average. Crucially, this pattern was similar for cSSIM which takes into account dynamic aspects of the prediction.

Finally, we inspected the prediction error (PE) across the PredNet model hierarchy (Fig. 4d). Similar to what we found for the above tested deep vision models, we expected that if PredNet extracted high-level representation, higher levels should show smaller errors compared to lower levels when comparing natural videos with their metameric counterparts. In contrast, we found that the prediction error of lowest level was most of the time lower than at all other levels for both original and STST videos. Also, the amount of prediction error across levels did not show a clear sorted pattern aligned with its hierarchical structure. Importantly, we also did not find a clear differences of the prediction error between original and STST videos across all levels. Together, these results offered insights into the representational capabilities of PredNet suggesting that this model is insensitive to the high-level information present in the input data.

## Discussion

We proposed a Spatiotemporal Style Transfer (STST) algorithm as a flexible framework for dynamic visual stimulus generation. STST empowers vision scientists to manipulate and synthesize video stimuli for vision research in both biological and artificial systems. We capitalized on recent work on image style transfer [25] and dynamic texture synthesis algorithms [44]. We extended them to work reliably in the context of natural videos by means of perceptual stabilization techniques. Our method is based on a two-stream architecture that performs an optimization on consecutive frames of a natural target video to generate model metamers [45], dynamic stimuli that are invariant to content and texture in both spatial and temporal domain.

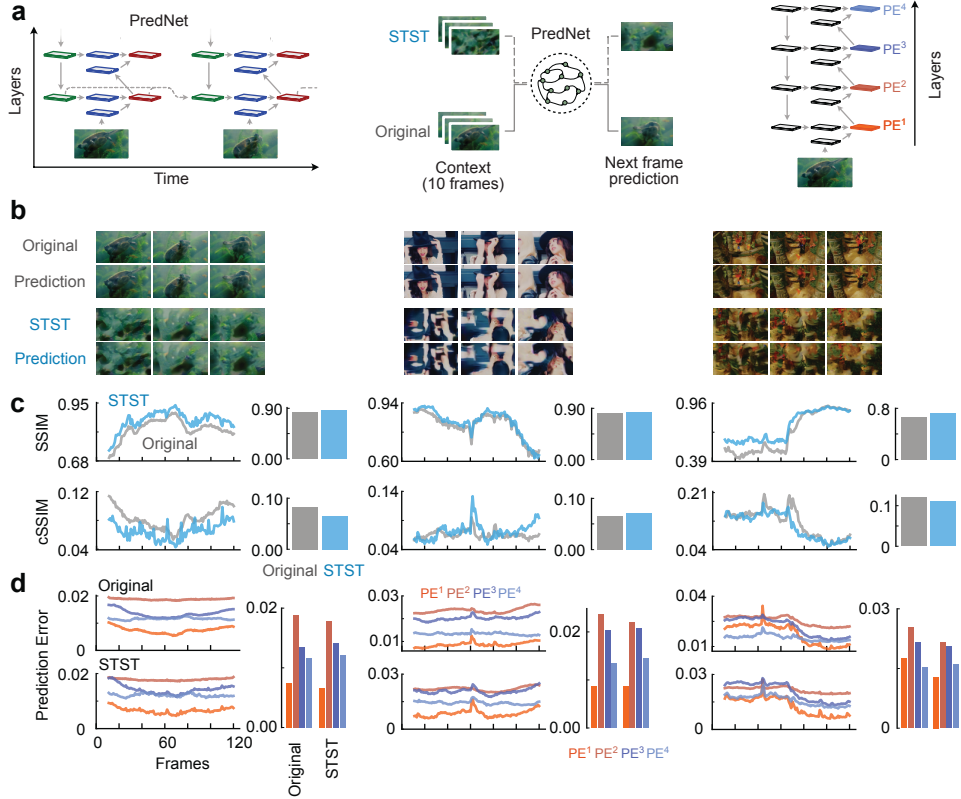

**Figure 4: Probing high-level information encoding in predictive coding networks using STST metamer stimuli. a**, Graphical outline of the analysis pipeline. Left: PredNet architecture with input and prediction (blue), representation (green) and error (red) modules. Gray arrows denote information flow, with solid lines indicating forward and feedback flow and the dashed lines indicating recurrent connections. Middle: We passed as inputs to PredNet generated STST stimuli and their natural counterparts (Original). We separately computed next frame predictions for both inputs. Right: We analyzed the predictive performance of the model and the prediction error (PE) representations of the error module across the four layers of hierarchy. **b**, Example ground truth and predicted frames. **c**, structural similarity index (SSIM, top) and conditional SSIM (bottom) across frames for original (gray) and STST (azure) stimuli. Barplots depict movie averages. **d**, prediction error across all 4 layers for original (top) and STST (bottom) stimuli. Barplots depict movie averages.

As a showcase example, we generated metamer stimuli that were invariant to the spatial and temporal texture of natural videos. The generated dynamic stimuli preserved important low-level spatiotemporal features of their natural counterparts and at the same time dismantle their high-level semantic information. Thus, these stimuli are suitable for investigating object recognition in visual systems. We showed that STST indeed preserves critical low-level visual features such as pixel intensity, contrast, global motion and optical flow. Moreover, we benchmarked our method to the only other, to the best of our knowledge, currently available method (STPS) [70] to generate dynamic stimuli preserving low-level spatiotemporal statistics. Crucially, STST outperformed STPS especially in preserving the optical flow of videos. This may be ascribed to the fact that STPS randomizes the phase of the spatiotemporal frequency spectrum, while our method uses as part of its objective the gram matrix of the layer activations coming from an optical flow estimation model. These findings show the potential to use our algorithm to study object recognition in dynamic natural vision.

We tested the effectiveness of STST generated stimuli in preserving low-level features and ablating high-level features using current state-of-the-art deep vision models [9,10]. These models parallel the hierarchical organization of the mammalian visual cortex, emulating the sequential extraction and integration of visual information from the distillation of low-level features in the initial layers to the more advanced layers that are responsible for the assimilation of high-level semantic content [54–56,71]. Notably, we employed convolutional-only models in our analysis since it has been shown that Vision Transformer architectures [85,86] do not exhibit the hierarchical structure in the visual feature extraction process [87]. Indeed, we found that, in image classification models, the early layers were similarly activated by the metamer and natural stimuli, confirming that they shared similar spatial low-level features. Crucially, we found that later stages in the model exhibited a decreased similarity indicating that late layer units selective for high-level features responded differently to natural and metamer stimuli. In addition, we observed the same pattern in video classification models. Importantly, these video processing models allowed us to confirm that early layers similarly encoded also low-level features in the temporal domain, since image processing models were applied to single frames. For the same logic, this analysis revealed that high-level information was disrupted in late layers also when integrating motion patterns. Notably, although our results on image processing models were expectedly revealing the hierarchical processing of visual features similar to the ventral stream in the visual cortex [56], there is less research on the effective nature of the hidden representations of video processing models and how they relate to biological visual systems [88,89].

Finally, we investigated the representational capabilities of predictive coding inspired deep neural networks performing next frame prediction [77–79]. We probed PredNet, a predictive coding recurrent neural network, with natural and metamer stimuli to investigate its predictive performance. This allowed us to test if PredNet only relies on low-level spatiotemporal features for next frame prediction or if instead it exhibits high-level representations of its inputs, as recently debated in the literature [77,83]. We found no evidence for PredNet to better predict next frames when probed with natural videos as compared to their STST counterparts, which suggests a lack of high-level semantic information in driving the information flow of the model. Importantly, the employed measure of predictiveness provided us a quantification of its predictive ability beyond the static image quality and the mere strategy of copying the previous frame as predictive for the next one [83], revealing its inability to better discriminate object-level motion patterns when available in the inputs as in natural videos and exploit them [77,83]. In addition, we also investigated the internal representations of PredNet in the error modules computing the prediction error between the top-down prediction and the bottom-up representation of the data across all levels of its hierarchical structure. Surprisingly, we found that the error values were not sorted according to the hierarchical structure as theoretically hypothesized in the predictive coding framework [90,91]. We speculate that this effect may be due to the fact that we used the version of PredNet trained to minimize only the lowest level prediction error, as it has been shown to performing best on next frame prediction tasks [79,83], while its architectural implementation differs from the theoretical proposition of predictive coding by predicting the prediction error at each layer instead of predicting the activation at the layer below [77]. These results offer critical insights into the representational capabilities of PredNet by leveraging our stimulus generation framework.

In conclusion, the proposed STST method will foster vision research using novel artificial dynamic stimuli. Our results shed light on potential applications of STST and promoted it as a versatile tool for dynamic stimulus generation.

## Acknowledgements

This study was supported by the European Research Council (ERC; https://erc. europa.eu/) CoG 864491 (M.S) and by the German Research Foundation (DFG; https://www.dfg.de/) projects 276693517 (SFB 1233) (M.S.) and SI 1332/6-1 (SPP 2041) (M.S.).

## Author contributions

AG: Conceptualization, Software, Methodology, Investigation, Formal analysis, Visualization, Writing – original draft, Writing – Review & Editing

MS: Conceptualization, Supervision, Resources, Project administration, Funding acquisition, Writing – Review & Editing

## Competing interest statement
The authors declare no competing interests.

## Materials and Methods

### Stimulus generation
We generated dynamic stimuli by matching spatiotemporal features to their natural counterparts. We optimized metameric stimuli to match both the spatial and temporal texture of 3 high-quality video clips collected from the YouTube-8M dataset [69]. The target videos consisted of 120 frames at 60 frames per second (FPS), with a frame resolution of 360×640. We included in our total loss function only the spatial $\mathcal{L}_{st}^{\ell}$ and temporal $\mathcal{L}_{tt}^{\ell}$ texture loss components to eliminate any high-order regularity. For the spatial stream $S(x_t)$, we used VGG-19 [57] and selected the layers conv1_1, conv2_1, conv3_1, conv4_1 and conv5_1 for the texture loss, while for the temporal stream $T(x_t, x_{t-1})$ we used MSOE [44] and selected the concatenation layer, where the multiscale feature maps of orientations are stored, for the texture loss. For each frame, we selected 3 octaves $o \in [-2, -1, 0]$ with an octave scale $\sigma = 1.5$, resulting in frame resolutions of 160×284, 240×426 and the original one. The weights associated to both texture losses ($\beta$ and $\lambda$) were always set to 1, while the weight $\omega$ for the $\mathcal{L}_{TV}$ loss was set dependently to the octaves, namely having a value of 0.05, 0.1 and 0.5, respectively. Moreover, the hyperparameters of the optimization process were octave-dependent, with the number of iterations being 250, 750 and 1000 and the learning rate $\eta$ set as 0.001, 0.003 and 0.005, respectively. We set the blending ratio $\varphi$ to 0.95 and the number of padding frames $\xi$ to 5.

### Comparison to existing methods
We generated dynamic metamer stimuli from the same target videos with a different algorithm [70], which we referred as Spatiotemporal Phase Scrambling (STSP). STSP consisted of three steps, starting with the randomization of the phase spectrum of each frame separately using a 2D fast Fourier transform (FFT). The random phase angles were sampled from a uniform distribution with range $[-\pi, \pi]$ and were then applied to all frames. Second, we used a 3D FFT to also randomize the phase spectrum of the full spatiotemporal data. Finally, we used the same color transfer algorithm [67] as for STST to enable a fair comparison of methods. We applied the first step to each channel of the frames separately, the second to all frames together but for each channel separately and the third step to each frame, since it handles the full 3D distribution of pixel intensity.

### Spatiotemporal feature analysis
We computed 4 basic spatiotemporal features from the original movies, our STST stimuli and the STPS stimuli. For the spatial low-level features, for each frame, we computed the average pixel value across all $H$, $W$ and $C$ dimensions and the luminance contrast as the standard deviation across both $H$ and $W$ [70], after transforming the image from the RGB color space to grayscale using the dot product between the channel values and the following vector [0.299, 0.587, 0.114]. For temporal low-level features, we computed the pixel change as the average across all $H$, $W$ and $C$ dimensions of the absolute difference between consecutive frames [70] and the optical flow feature as the average across $H$ and $W$ of the magnitude and the angle of the dense optical flow estimation between consecutive frames (grayscaled as above) using the Farnebäck algorithm [92] with pyramid scale set as 0.5 with 5 levels, an averaging window size of 13, 10 iterations per level, 5 pixel neighbours used to find polynomial expansion in each pixel and a Gaussian kernel with a standard deviation of 1.1 for smoothing the derivatives used as a basis for the polynomial expansion.

### Testing vision models on metamer stimuli
We tested the effects of STST stimuli on hidden activations of current state of the art deep learning models for image and video classification. We used ResNet50 [73] and ConvNeXt-T [74] trained on the Imagenet1K dataset [59] as image classification models and ResNet18-3D [75] and ResNet18-MC3 [76] trained on Kinetics400 dataset [93] as video classification models. For video classification models, we used 5 consecutive frames as input and progressed throughout the video with a 1 frame step size. For image classification models, we processed individual frames. As early layers, we selected the "maxpool" layer for ResNet50, the "features.0" for ConvNeXt-T and the "layer1.0.conv1.1" for both ResNet18-3D and ResNet18-MC3. As late layers, we selected the "layer4.2.relu_2" layer for ResNet50, the "features.7.2.add" for ConvNeXt-T and the "layer4.1.relu" for both ResNet18-3D and ResNet18-MC3. This layer nomenclature was adopted from the *torchvision* [94] implementation of these models. Center Kernel Alignment (CKA) [72] with a linear kernel was used as a similarity score of the models' layer activations between natural and STST stimuli.

### Testing high-level representations in predictive coding networks
We used metamer stimuli to investigate the role of high-level representations in PredNet [79], a deep convolutional recurrent neural network with 4 hierarchical levels that was trained with self-supervised learning methods with the KITTI dataset [95] to perform next frame prediction [80]. We used the same implementation of the original work from Lotter et al. [79] and focused our analyses on the model which had the highest predictive performance [79,83], namely the one that was trained to minimize the prediction error from the lowest level of the hierarchical structure of PredNet (in the original paper referred to as PredNet $L_0$). We passed to the model the metamer stimuli alongside their natural counterparts in a sliding-window fashion, using as context the last 10 frames, in line with the original training hyperparameter [79]. The inputs were preprocessed by rescaling them to a frame resolution of 128×160 as required by the model. We extracted from the model the predicted next frame and the average prediction error coming from the output of the error module across all levels of the hierarchy. We then computed the structural similarity index measure (SSIM) [84] and its dynamic version conditional on the previous frame (cSSIM) [83]. cSSIM measures how different the predictions are from the previous frame, quantifying how risky is the prediction of the model in comparison to simply performing a copying the previous frame as a prediction for the current one:

$$cSSIM(f_t, f_{t-1}, \hat{f}_t) = \left(1 - SSIM(f_{t-1}, \hat{f}_t)\right) \cdot SSIM(f_t, \hat{f}_t) \qquad (10)$$

where $f_t$ denotes the current frame (either from the original or STST video) and $\hat{f}_t$ represents PredNet's prediction of the current frame.